%% file: CVPR.tex
\documentclass[10pt,twocolumn,letterpaper]{article}

\usepackage[pagenumbers]{cvpr} 

\input{preamble}

\definecolor{cvprblue}{rgb}{0.21,0.49,0.74}
\usepackage[pagebackref,breaklinks,colorlinks,citecolor=cvprblue]{hyperref}

\usepackage{multirow}
\usepackage{colortbl}
\usepackage{wrapfig}
\usepackage{pifont}

\def\paperID{*****} 
\def\confName{CVPR}
\def\confYear{2024}

\title{Joint Point Cloud Upsampling and Cleaning with Octree-based CNNs}

\author{
  Jihe Li\quad \quad
  Bo Pang\quad \quad
  Peng-Shuai Wang\thanks{Corresponding author.} \\
  Peking University, Beijing, China \\
  \small
  \texttt{lijh@stu.pku.edu.cn, bbouc98@gmail.com, wangps@hotmail.com}
}

\begin{document}
\maketitle

\input{src/abstract.tex}
\input{src/introduction}

\input{src/relatedworks}
\input{src/method}

\input{src/results}

\input{src/conclusion}

\input{src/acknowledgements}

{
    \small
    \bibliographystyle{ieeenat_fullname}
    \bibliography{refs}
}


\end{document}

%% file: preamble.tex
\usepackage[dvipsnames]{xcolor}


%% file: src/abstract.tex
\begin{abstract}
Recovering dense and uniformly distributed point clouds from sparse or noisy data remains a significant challenge.
Recently, great progress has been made on these tasks, but usually at the cost of increasingly intricate modules or complicated network architectures, leading to long inference time and huge resource consumption.
Instead, we embrace simplicity and present a simple yet efficient method for jointly upsampling and cleaning point clouds.
Our method leverages an off-the-shelf octree-based 3D U-Net (OUNet) with minor modifications, enabling the upsampling and cleaning tasks within a single network.
Our network directly processes each input point cloud as a whole instead of processing each point cloud patch as in previous works, which significantly eases the implementation and brings at least 47 times faster inference.
Extensive experiments demonstrate that our method achieves state-of-the-art performances under huge efficiency advantages on a series of benchmarks.
We expect our method to serve simple baselines and inspire researchers to rethink the method design on point cloud upsampling and cleaning.
Our code and trained model are available at \url{https://github.com/octree-nn/upsample-clean}.
\end{abstract}

%% file: src/introduction.tex
\begin{figure*}[htbp]
    \centering
   \begin{subfigure}[b]{0.32\linewidth}
       \centering
       \includegraphics[width=0.9\linewidth]{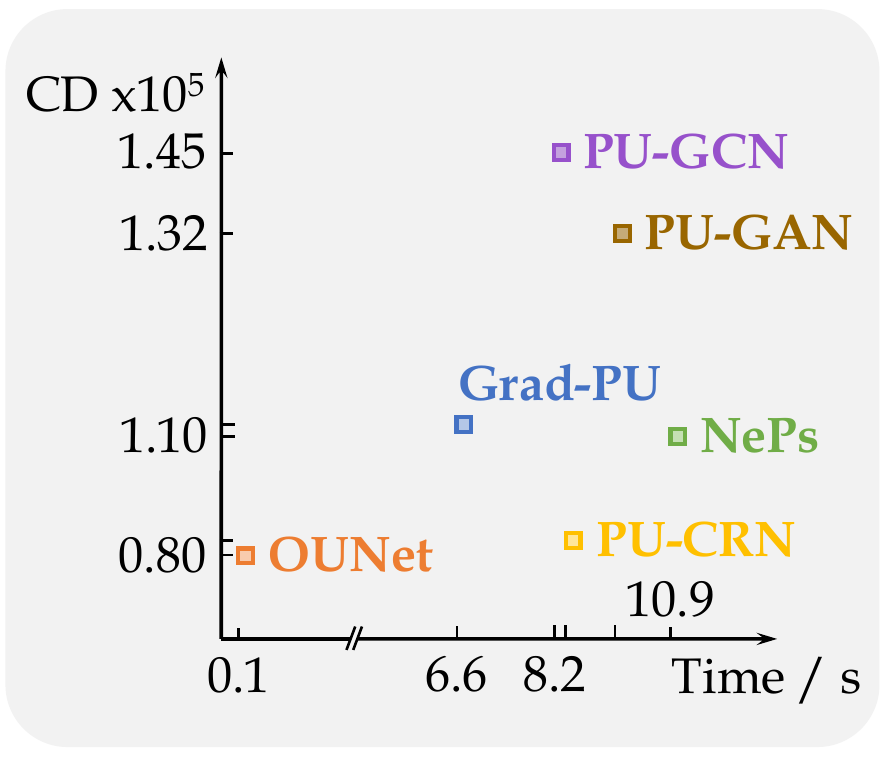}
       \subcaption{Upsampling}
       \label{fig:speed-up}
   \end{subfigure}
   \centering
   \begin{subfigure}[b]{0.32\linewidth}
       \centering
       \includegraphics[width=0.9\linewidth]{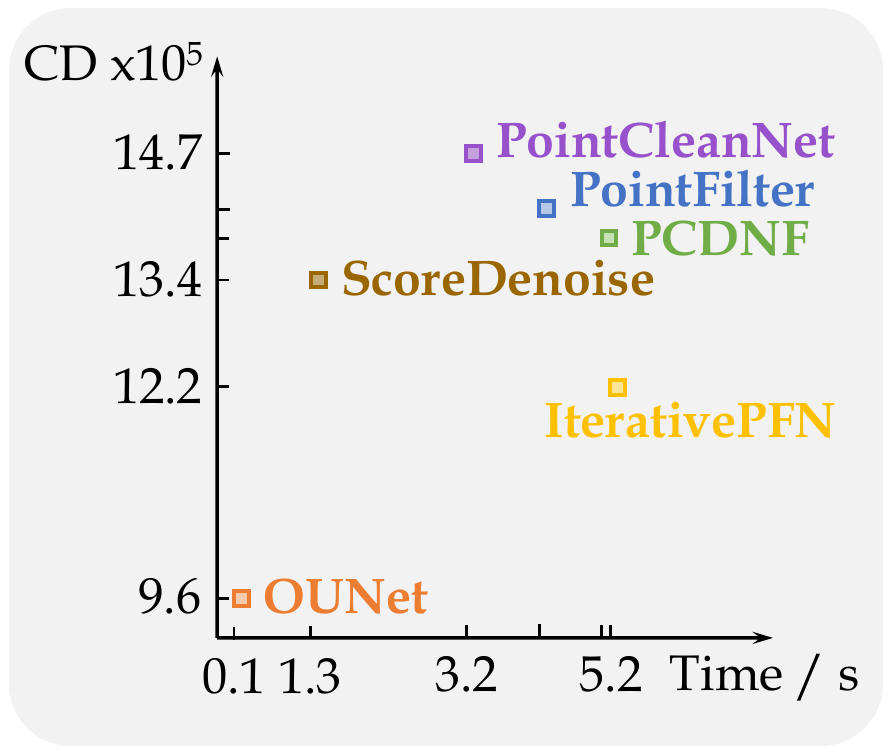}
       \subcaption{Cleaning (10k points)}
       \label{fig:speed-10}
   \end{subfigure}
   \centering
   \begin{subfigure}[b]{0.34\linewidth}
       \centering
       \includegraphics[width=0.9\linewidth]{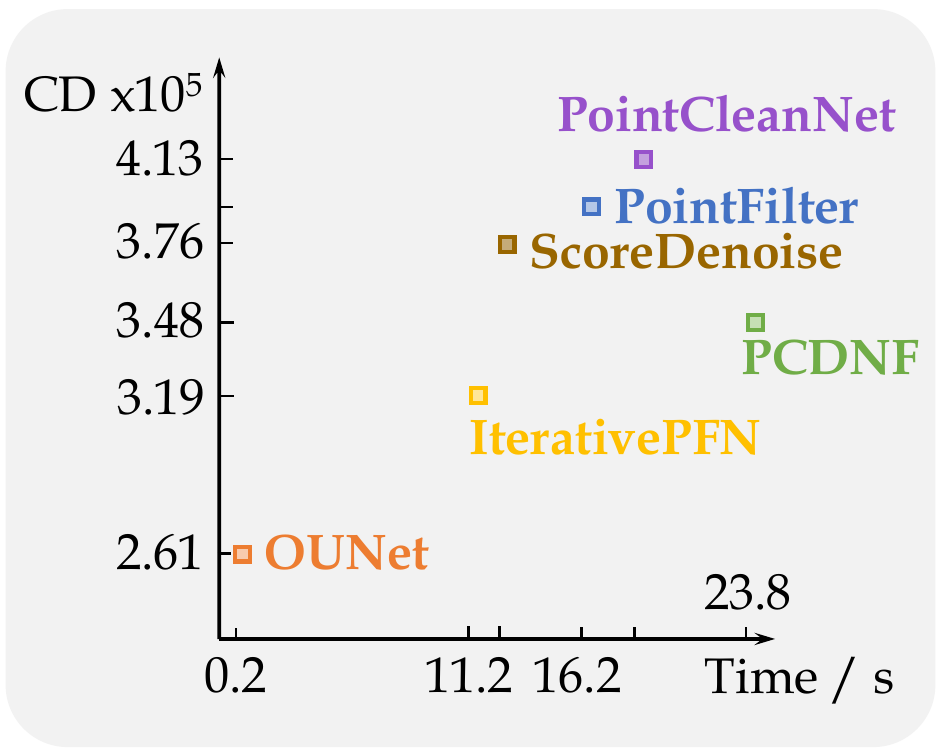}
       \subcaption{Cleaning (50k points)}
       \label{fig:speed-50}
   \end{subfigure}
    \caption{Comparisons of our method (OUNet) with other state-of-the-art methods. The horizontal and vertical axes represent the running time of generating an entire point cloud and the Chamfer distance, respectively. (a): The results of upsampling on the PU-GAN dataset from 5k to 80k points. (b): The results of cleaning 10k points with noise level 0.02. (c): The results of cleaning 50k points with noise level 0.02. Our method achieves the best Chamfer distances while running the fastest. It runs at least \emph{47 times} faster than the others on point upsampling.}
   \label{fig:teaser}
\end{figure*}

\section{Introduction}
With advancements in 3D acquisition devices, point cloud data has seen widespread applications in robotics, autonomous driving, and 3D modeling~\cite{li2020deep, song2021vis2mesh, chen20203d}. However, real-world point cloud data often suffers from sparsity and noise due to complex environments and inherent limitations of 3D scanners. Therefore, recovering dense and uniform point clouds becomes crucial for various 3D tasks like classification, segmentation, and reconstruction~\cite{xu2020geometry, liu2021pointguard, fan2021scf, huang2023learning}.

The inherently sparse and unstructured nature of point clouds presents a significant challenge for upsampling and cleaning tasks, particularly when aiming to recover fine-grained details and achieve uniform distribution.
Traditional optimization-based methods~\cite{alexa2003computing,guennebaud2007algebraic,lipman2007parameterization,preiner2014continuous,wu2015deep,digne2017bilateral,sun2015denoising,huang2013edge,dinesh20193d} rely heavily on human priors, often struggling to work well in challenging scenarios. 
Recently, learning-based methods have shown great potential in these tasks and demonstrated increasing effectiveness.
However, their network architectures are getting more complicated.
For instance, CPU~\cite{zhao2023cpu} divided the upsampling pipeline into three stages, including training a denoising autoencoder, optimizing a codebook, and predicting dense codes.
IterativePFN~\cite{de2023iterativepfn}, designed for point cloud denoising, stacked a sequence of encoder-decoder pairs and uses multi-scale noisy point clouds to supervise the intermediate outputs.
Longer pipelines and more complex architectures often lead to increased computational costs and training difficulties.

Additionally, most existing methods rely on intricate pre- and post-processing steps, like farthest point sampling (FPS) or distance weight based stitching scheme~\cite{de2023iterativepfn}.
Furthermore, while point cloud upsampling and cleaning are similar tasks, many existing methods still address upsampling or cleaning independently~\cite{wei2021geodualcnn, liu2023pcdnf, de2023iterativepfn, wei2024pathnet}, making it challenging to leverage the shared features between the two tasks.

In this paper, we present a simple yet effective solution with a single neural network for both point cloud upsampling and cleaning.
We represent the point cloud with a multi-scale octree structure~\cite{wang2017cnn}, enabling the processing of varying numbers of points and the extraction of both global and fine-grained features without splitting the point cloud into patches.
This simple and straightforward strategy significantly simplifies processing procedures, reduces inference time, and facilitates joint training.
We leverage an off-the-shelf octree-based U-Net network~\cite{wang2020deep,wang2017cnn}, incorporating a key modification to achieve joint training: replacing the batch normalization in \cite{wang2020deep}  with the group normalization.
This change prevents the mixing of upsampling and cleaning features, thereby maintaining high performance.
Extensive experiments and comparisons demonstrate that our method achieves state-of-the-art results on metrics such as Chamfer Distance (CD), Hausdorff Distance (HD), and Point-to-Surface Distance (P2F).

We intentionally avoid introducing superfluous modules, which leads to a straightforward solution and is easy to understand and implement.
Extensive experiments verify that our straightforward method performs even better than those with complicated designs.
The speed and performance comparisons are shown in Fig.~\ref{fig:teaser}.
Compared with other state-of-the-art methods, our method achieves the best Chamfer distances while running at least 47 times faster than the others on point upsampling.
We expect our method to serve as simple baselines that unify point cloud upsampling and cleaning into a single framework.
Our work could potentially inspire the community to rethink the development of research on point cloud upsampling and cleaning.

%% file: src/relatedworks.tex
\section{Related Works}
In this section, we mainly review learning-based methods for point cloud upsampling and cleaning.
We categorize existing methods into groups based on their key ideas and strategies.
We split the related works into two parts: point cloud upsampling and point cloud cleaning, as most methods are designed for only one of these tasks.

\subsection{Point Cloud Upsampling}
\paragraph{Displacement-based Methods.}

These methods typically predicted multiple displacements for each point in a sparse point cloud~\cite{yu2018pu, li2021point, qian2021pu, qian2021deep, luo2021pu, ye2021meta, qiu2022pu, du2022point}. PU-Net~\cite{yu2018pu} presented the first point cloud upsampling method based on neural networks in a per-patch style. Its key idea is to learn point-wise features and split each one into multitude in the latent space, which are used to predict displacements. Dis-PU~\cite{li2021point} and PU-CRN~\cite{du2022point} employed cascaded subnetworks to split the total displacements prediction into multiple stages. PU-GCN~\cite{qian2021pu} and PU-Transformer~\cite{qiu2022pu} introduced graph convolution and Transformer architectures to extract point features, respectively.
PU-GAN~\cite{li2019pu} employed a compound loss instead of only an adversarial loss.
PUFA-GAN~\cite{liu2022pufa} designed a frequency-aware discriminator which consists of global and high-frequency sub-discriminators. Meta-PU~\cite{ye2021meta} employed a meta-subnetwork to take the scale factor as input and predicted the weight of the Meta-RGC block for feature extraction.

\paragraph{Parametric Sampling Methods.}
Several methods~\cite{qian2020pugeo,feng2022neural,wei2023ipunet} learned parametric surfaces to pull the sampled points from a canonical space to the underlying surface. For instance, PUGeo-Net~\cite{qian2020pugeo} sampled new points in a 2D domain, computed normals, mapped them to the tangent plane, and then projected them to the 3D surface by adding learned displacements along the normals. Different from PUGeo-Net, iPUNet~\cite{wei2023ipunet} incorporated a point-based cross-field to build the tangent plane and generated new points on the tangent plane directly.

\paragraph{Iterative Methods.}
An iterative strategy has been widely employed to optimize the position of points progressively~\cite{yifan2019patch,kumbar2023tp,wei2023ipunet,he2023grad}. MPU~\cite{yifan2019patch} cascaded multiple networks, each assigned to upsample the point cloud with a small ratio. In each iteration, TP-NoDe~\cite{kumbar2023tp} concatenated a noise perturbation with a sparse point cloud patch, followed by a network to denoise it and output an upsampled version.

Unlike the various methods mentioned above, we discarded the patch-style manner, iterative strategy, and parametric surfaces, and instead employ a straightforward U-Net architecture to upsample the point cloud end-to-end.

\subsection{Point Cloud Cleaning} \label{sec:relatedworks_cleaning}

\paragraph{Displacement-based Methods.} For point cloud cleaning, a common approach is also to extract features and predict displacements to the target surface for each point~\cite{pistilli2020learning, mao2022pd, chen2022repcd, wang2023random, wei2024pathnet}. GPDNet~\cite{pistilli2020learning} employed DGCNN~\cite{simonovsky2017dynamic} to extract point-wise features for regressing residual displacements.
RePCD-Net~\cite{chen2022repcd} designed a feature-aware recurrent network that aggregates multi-scale features to enhance geometric perception across stages. PathNet~\cite{wei2024pathnet} introduced a multi-path denoising method based on Reinforcement Learning (RL) to dynamically select the most suitable path for each point. 

\paragraph{Normal Integration Methods.}
Many methods integrated normals to provide more geometric information than just the positions of points. PointProNets~\cite{roveri2018pointpronets} projected a 3D point cloud patch into a 2D heightmap along the normal of its center point, followed by a Convolutional Neural Network to filter noise. Several methods~\cite{luo2020differentiable, lu2020deep, wei2021geodualcnn, liu2023pcdnf} incorporated normals of point clouds to construct potential manifolds or enhance geometric priors. GeoDualCNN~\cite{wei2021geodualcnn} first predicted the normals of noise-free point clouds via two parallel networks, and then updated the positions of points along their normals by minimizing three quadrics. PCDNF~\cite{liu2023pcdnf} incorporated a normal filtering task to enhance the cleaning ability of the network while preserving geometric details.
ResGEM~\cite{zhou2024resgem} adopted dual graph convolution branches for normal regression and vertex offset estimation to strike a balance between smoothness and geometric details.

\paragraph{Iterative Methods.} Similar to point cloud upsampling, many point cloud cleaning methods~\cite{rakotosaona2020pointcleannet, zhang2020pointfilter, luo2021score, chen2022deep, de2023iterativepfn} also employed iterative strategies. Pointfilter~\cite{zhang2020pointfilter} took a patch of noisy point cloud and predicted the displacement to the underlying surface of the center point in each iteration. ScoreDenoise~\cite{luo2021score} updated the coordinates of points iteratively by gradient ascent under the guidance of a trained score estimation network. Furthermore, PSR~\cite{chen2022deep} designed a kernel-based smoothing module in the network and regularization during the optimization stage to enhance continuity.

In this paper, we adopted an identical network for both point cloud upsampling and cleaning, which significantly simplifies the network design and implementation.

%% file: src/method.tex
\section{Approach}
In this section, we first review previous methods for point cloud upsampling and cleaning and analyze their limitations (Sec.~\ref{sec:review}). Subsequently, we elaborate on our approach for reconstructing high-resolution and clean point clouds through the joint training of a model that operates on the entire point cloud using an octree-based network (Sec.~\ref{sec:network}).

\subsection{Analysis} \label{sec:review}

\begin{figure}[t]
    \centering
    \begin{subfigure}[b]{0.25\textwidth}
        \centering
        \hspace*{-8pt}
        \includegraphics[width=0.99\textwidth]{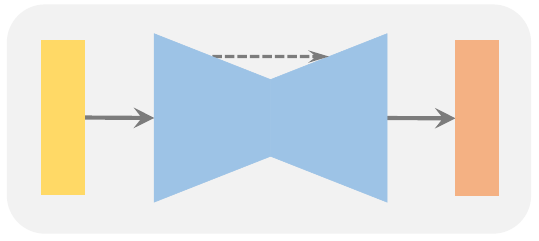}
        \caption{OUNet}
        \label{fig:ounet}
    \end{subfigure}
    \hfill
    \begin{subfigure}[b]{0.49\textwidth}
        \centering
        \hspace*{-8pt}
        \includegraphics[width=0.99\textwidth]{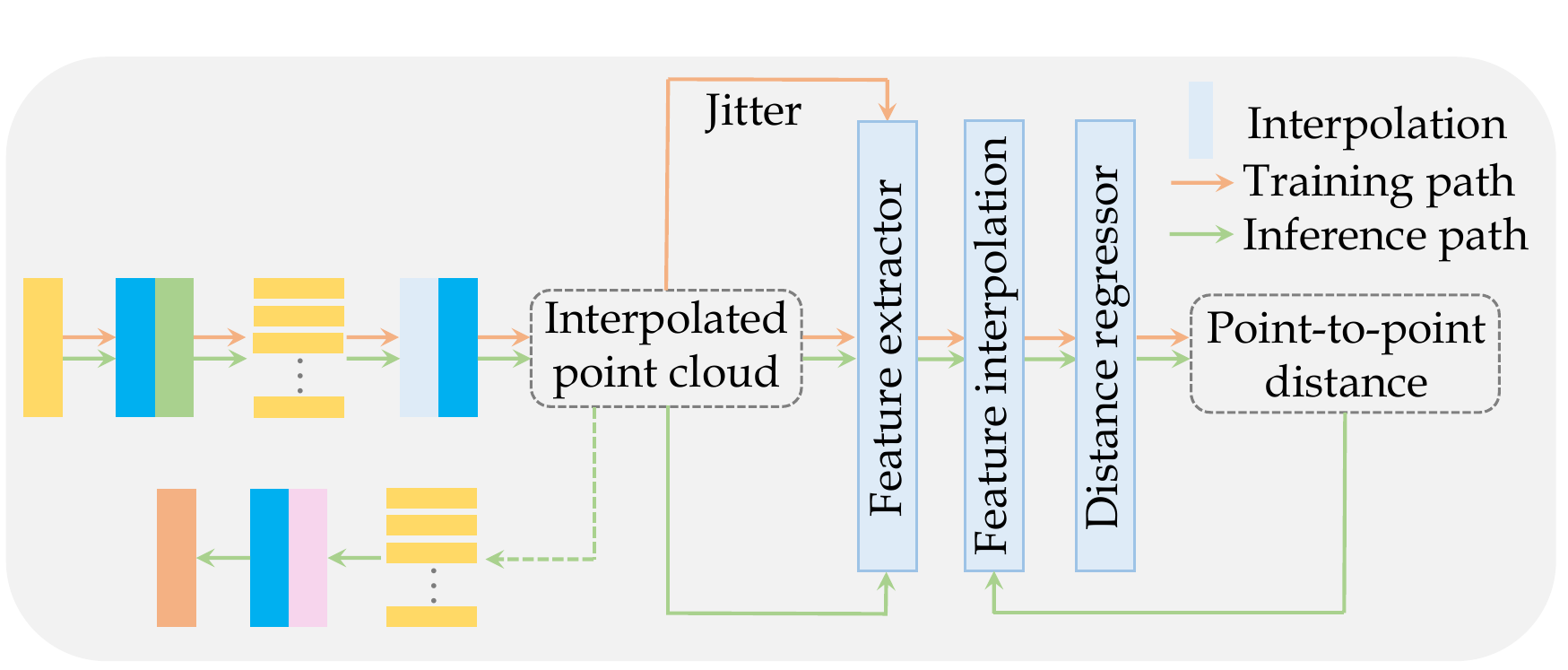}
        \caption{Grad-PU}
        \label{fig:gradpu}
    \end{subfigure}
    \hfill
    \begin{subfigure}[b]{0.49\textwidth}
        \centering
        \hspace*{-8pt}
        \includegraphics[width=0.99\textwidth]{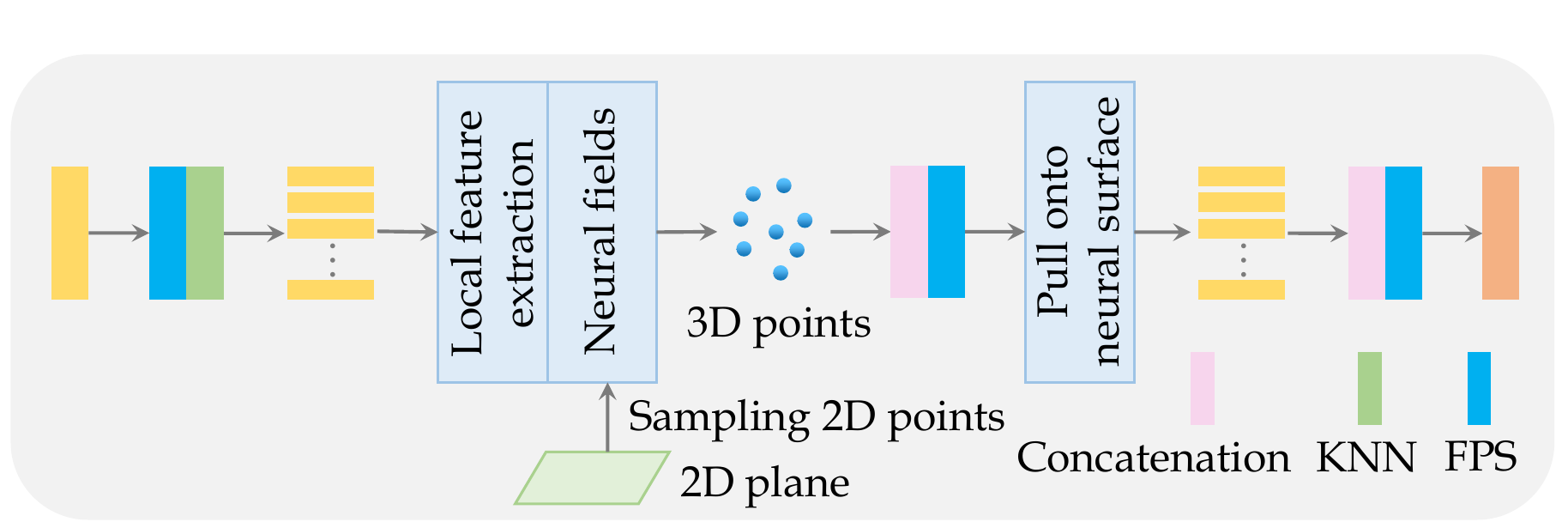}
        \caption{Neural Points}
        \label{fig:neural}
    \end{subfigure}
    \hfill
    \begin{subfigure}[b]{0.49\textwidth}
        \centering
        \hspace*{-8pt} 
        \includegraphics[width=0.99\textwidth]{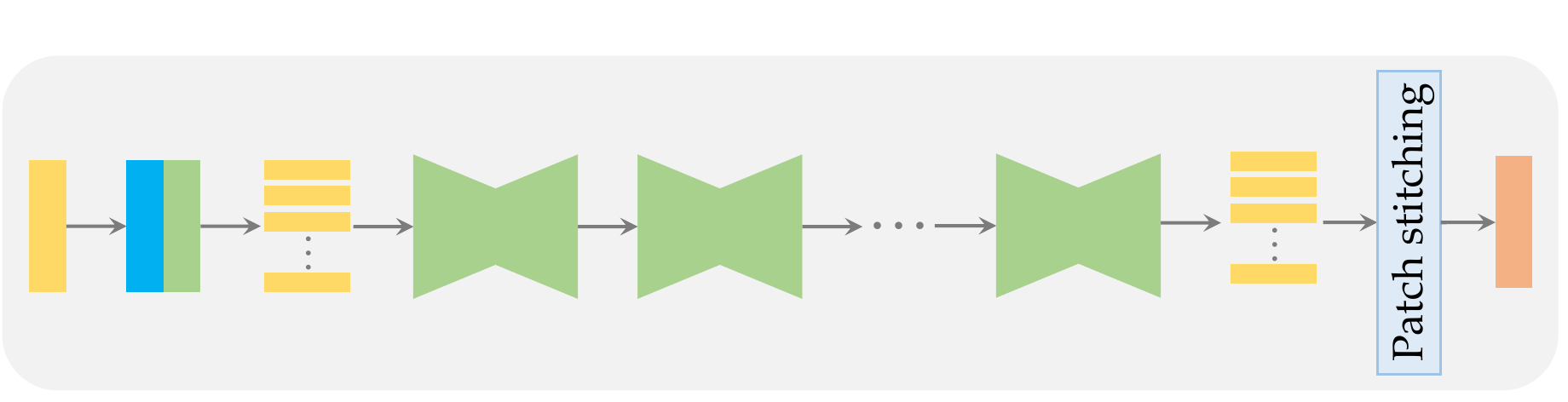}
        \caption{IterativePFN}
        \label{fig:iterativepfn}
    \end{subfigure}
    \caption{Pipeline comparisons between our OUNet and other methods on point cloud upsampling and cleaning. (a): OUNet. The model consists of multiple octree-based Residual Blocks, Downsampling, and Upsampling components. (b)-(d): These 3 methods all require splitting point cloud into patches using FPS and K-Nearest Neighbors (K-NN) and upsampling or cleaning them, followed by stitching patches. In this figure, all icons sharing the same shape and color convey identical meanings. The \emph{yellow} and \emph{orange} rectangles denote input and output point cloud, respectively.}
    \label{fig:methods}
\end{figure}

In Fig.~\ref{fig:methods}, we compare our pipeline with existing methods. As illustrated in Fig.~\ref{fig:ounet}, our approach uses a straightforward 3D U-Net to process the entire point cloud, significantly streamlining the procedure. In contrast, methods like Grad-PU~\cite{he2023grad}, Neural Points~\cite{feng2022neural}, and IterativePFN involve more complex data processing and architectures.

These state-of-the-art methods typically require dividing the point cloud into patches to extract geometric details before feeding them into the network and subsequently have to stitch these patches together to reconstruct the entire point cloud.
The pre- and post-processing, such as FPS, cannot be parallelized on GPUs. They tend to introduce implementation complexity, long inference times, and computational burdens. Additionally, the patches often contain significant overlap to prevent holes when stitched together, also leading to substantial redundant computation.

For instance, Fig.~\ref{fig:gradpu} depicts the network architecture of Grad-PU. It first employs traditional interpolation and FPS algorithms to obtain a target resolution point cloud, followed by training a Distance regressor to estimate the distance between the input and target point clouds. During inference, it iteratively optimizes the positions of the interpolated point cloud to minimize the estimated distance, making the process more time-consuming and complex compared to our method.

Neural Points, achieving state-of-the-art performance on the Sketchfab benchmark, constructs a local Neural Field for every point, mapping new 2D samples to 3D space, and then projects the 3D points onto a neural surface to improve smoothness, as shown in Fig.~\ref{fig:neural}.

IterativePFN stacks a series of 3D autoencoders, as depicted in Fig.~\ref{fig:iterativepfn}, and requires less noisy point clouds to supervise the first and intermediate modules, significantly slowing down the network's forward and backward passes. In contrast, our method employs a single 3D U-Net, which is much simpler and faster than IterativePFN.

\subsection{Network} \label{sec:network}
\begin{figure}[htbp]
    \centering
    \includegraphics[width=0.98\linewidth]{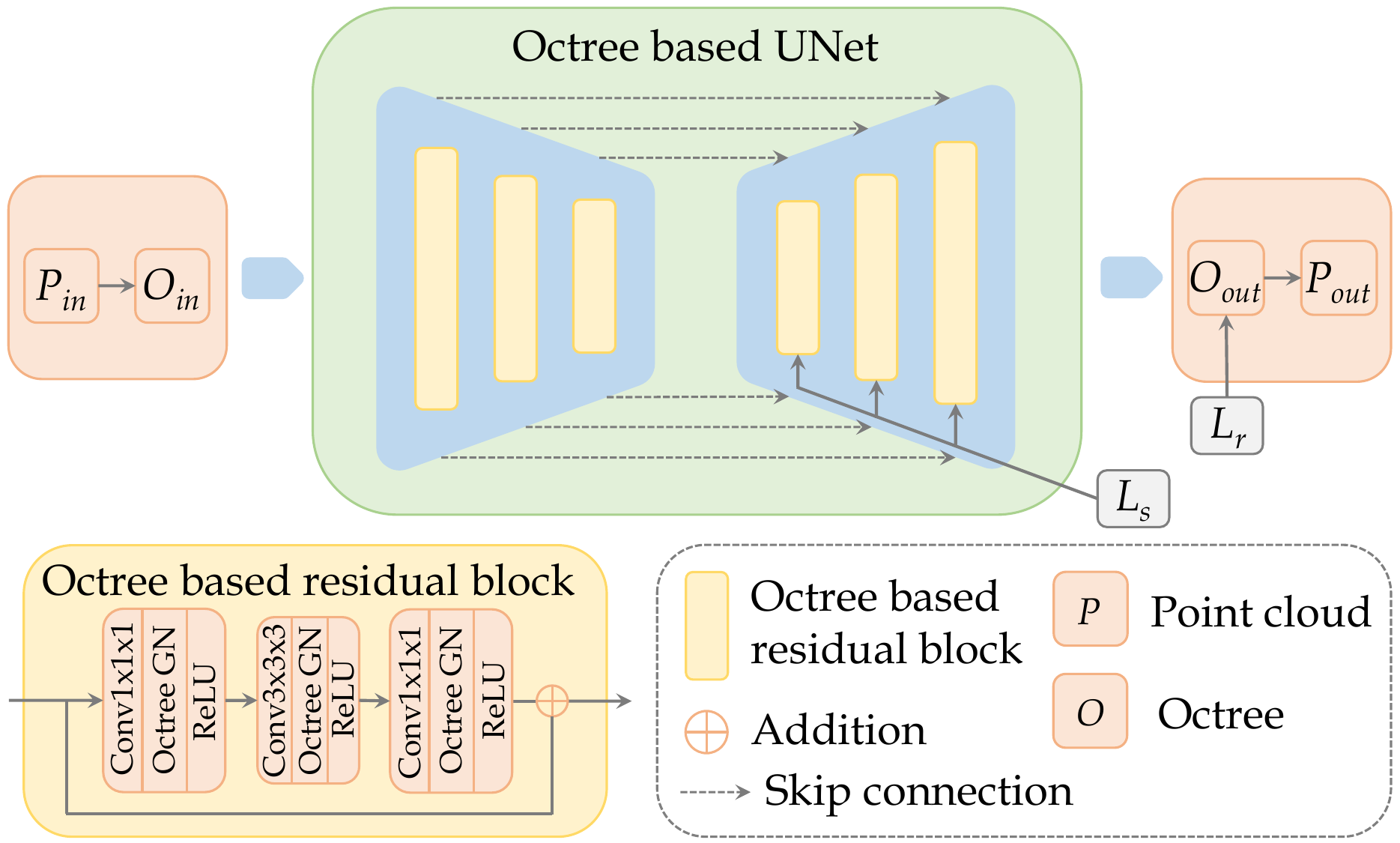}
    \caption{Architecture of OUNet. We represent the input point cloud as octrees.
    The architecture is simple, consisting of the encoder, decoder and skip connections. Both the encoder and decoder employ stacked octree-based residual blocks.}
    \label{fig:ounet-detail}
\end{figure}

The detailed architecture of our method is illustrated in Fig.~\ref{fig:ounet-detail}. Compared with previous methods, our network not only employs a much simpler and faster octree-based U-Net architecture, but also can jointly train a single network for both point cloud upsampling and cleaning.

\paragraph{Input and Output.}
The input of our network is a point cloud, which is then transformed into an octree structure.
The output is a high-resolution point cloud or a clean version of the input point cloud, which is derived from a processed octree.
Octree structure allows our network to handle point clouds with varying resolutions without modifying hyperparameters. This way, sparse or noisy point clouds can be fed into the network in one batch, enabling joint training for upsampling and cleaning. Additionally, the hierarchical octree helps preserve and reconstruct fine-grained geometry, allowing our network to process the entire point cloud without splitting it into patches, avoiding time-consuming pre- and post-processing.

\paragraph{Detailed Architecture.}
As shown in Fig.~\ref{fig:ounet-detail}, our network consists of an O-CNN encoder and decoder with output-guided skip connections~\cite{wang2017cnn,wang2020deep}. The encoder takes the octree built from the point cloud as input and constrains the convolution computation by only executing octree nodes. Through a series of 3D convolution operations, features extracted from input point coordinates are processed and downsampled along the octree structure in a bottom-up manner to generate hierarchical feature maps.
Unlike previous patch-based methods, our approach can extract global features, which are crucial for maintaining geometric consistency across the entire point cloud. In the decoder, the target octree is initialized as a low-resolution full octree and grows in a top-down order. The coarse-to-fine prediction scheme enhances detail preservation and overall reconstruction quality, while being computationally efficient and scalable. \looseness=-1

At each level of the octree, a shared prediction module determines whether a node is empty or not. Non-empty nodes are split into eight octants, and the feature of each node is propagated to its children. Repeat the process above until the specified octree depth is reached. At maximum depth, the displacement are predicted at each non-empty node to describe the final shape.

Notably, we employ octree-based group normalization (GN) in convolutional layers instead of batch normalization (BN), which is crucial for jointly implementing point cloud upsampling and cleaning. BN causes mixing of upsampling and cleaning features, while GN ensures their independence. In Sec.~\ref{sec:ablation}, we demonstrate the effectiveness of this modification through contrast experiments.
Also, experimental results in Sec.~\ref{sec:experiments} suggest that such simple modifications can significantly improve the performance of our network. \looseness=-1

The skip connection links the encoder and decoder at each level, retaining information from the encoder. Due to the asymmetry between the encoder and decoder using sparse convolution, feature maps from the encoder cannot be directly added via traditional skip connections. Additionally, features of noisy octree nodes should not be retained. Therefore, we adopt output-guided skip connections~\cite{wang2020deep} to retain useful information, following this rule: the skip connection between the encoder and decoder is added only for a non-empty octree node in the output when there is an input octree node in the same location.

\paragraph{Loss.}
For optimization, we define two types of loss: \textit{Structure loss} and \textit{Regression loss}, denoted as $L_{s}^l$ and $L_{r}$, respectively. Structure loss is the cross-entropy loss between the predicted \textit{status of nodes} and the ground-truth counterpart at each level of octree. Regression loss is the Mean Squared Error (MSE) of \textit{local coordinates}. Therefore, the total loss function of the network is defined as
\begin{equation}
\label{eq:loss}
    Loss = \sum_{l=2}^dL_{s}^l + L_{r},
\end{equation}
where $d$  is the maximum octree depth.

%% file: src/results.tex
\section{Experiments} \label{sec:experiments}
In this section, we first introduced our experiment configurations and several baseline methods in Sec.~\ref{sec:metrics} to \ref{sec:comparisons}.
Then, we presented the detailed results of both our method and the baseline methods in Sec.~\ref{sec:results}.
Finally, we conducted ablation studies to analyze each design of our method in Sec.~\ref{sec:ablation}.

\subsection{Evaluation Metrics} \label{sec:metrics}
Following the majority of existing methods~\cite{feng2022neural,he2023grad,qian2021pu,du2022point},
we adopted three metrics to evaluate the performance of our model: CD, HD, and P2F.
Each metric reflecte a different aspect of quality. CD measures the average distance between two point sets, providing an overall assessment of their alignment.
HD is sensitive to outliers and holes, making it useful for identifying regions of significant misalignment.
P2F measures the minimum distance from each point in one set to the surface of the other, providing a more local assessment of their proximity.

\subsection{Dataset} \label{sec:dataset}
\paragraph{Point Cloud Upsampling.}
In our experiments, we evaluated our model for point cloud upsampling on three widely used datasets: PU-GAN, Sketchfab, and PU1K~\cite{qian2021pu}. PU-GAN comprises 147 3D models from the publicly available PU-Net and MPU datasets. It containes 120 models for training and a separate set for testing, encompassing a diverse range of smooth and highly detailed objects.
Sketchfab consistes of 103 high-resolution 3D models, offering greater complexity compared to PU-GAN.
PU1K containes 1,000 models across 50 categories from ShapeNetCore~\cite{chang2015shapenet}. It splits the data into 900 models for training and the remaining for testing.
For both training and testing, we employed consistent preprocessing for the three datasets.
For each mesh, we generated an input point cloud with 5,000 points using Poisson Disk Sampling from the original mesh. Each ground-truth point cloud contains 500,000 points, obtained from Uniform Sampling. Our method necessitates sampling an even larger number of points for the ground truth due to the inevitable downsampling when building octree from point cloud.
Finally, all point clouds were normalized to the unit cube $[-1, 1]^3$.
Our data augmentation strategy includes random mirroring and random elastic deformations, as described in ~\cite{choy20194d}.

\paragraph{Point Cloud Cleaning.}
Following the protocols established in ScoreDenoise, PSR, and IterativePFN, we utilized the synthetic PU-Net and real-scanned Kinect v1 and v2~\cite{wang2016mesh} datasets to evaluate our model on point cloud cleaning. PUNet consists of 40 meshes for training and 20 meshes for testing.
We employed Poisson disk sampling to generate ground-truth point clouds containing 50,000 and 300,000 points for training the point cloud cleaning models at resolutions of 10,000 and 50,000 points, respectively.
Subsequently, the noisy point clouds were obtained by adding Gaussian noise with standard deviations ranging from 1$\%$ to 2$\%$ of the bounding sphere's radius.
During testing, we sampled 10,000 and 50,000 point cloud resolutions from the 20 testing meshes and introduced Gaussian noise with standard deviations of 1$\%$, 2$\%$ and 2.5$\%$ of the bounding sphere's radius.
We exclusively utilized the test sets of Kinect v1 and v2, which consist of 73 and 72 real-world scans, respectively, captured by Microsoft Kinect v1 and v2 cameras.
The data normalization and augmentation techniques employed in these experiments are identical to those presented in point cloud upsampling experiments.

\subsection{Implementation details} \label{sec:implementation}
We combined all the datasets mentioned in Sec.~\ref{sec:dataset} to train a single network jointly.
The octree was configured with a full level of 2 and a maximum level of 8. The channels of network were set to [256, 256, 128, 128, 64, 64, 32] from level 2 to level 8.
We employed the AdamW optimizer~\cite{loshchilov2018decoupled} for training, with an initial learning rate of 0.0005, a weight decay of 0.05, and a batch size of 8. We trained the network for 500 epochs and decayed the learning rate using a polynomial function with a power of 0.9.
All experiments were conducted on Nvidia RTX 3090 GPUs.

\begin{table*}[htbp]
    \centering
    \setlength{\tabcolsep}{9.9pt}
    \caption{Quantitative comparisons for point cloud upsampling on PU-GAN and Sketchfab and PU1K dataset in terms of CD ($\times$$10^{-5}$), HD ($\times$$10^{-4}$), P2F ($\times$$10^{-3}$), lower is better. Time is measured in seconds. Best results are highlighted as \colorbox{red!30}{first}, \colorbox{orange!30}{second}.}
    \begin{tabular}{l|rrr|rrr|rrr|r}
    \toprule
    \multirow{2}{*}{Models} & \multicolumn{3}{c|}{PU-GAN} & \multicolumn{3}{c|}{Sketchfab} & \multicolumn{3}{c|}{PU1K} & \multirow{2}{*}{Time} \\
    \cmidrule{2-10}
       &  CD   & HD   &  P2F  &  CD  &  HD  &  P2F  &  CD  &  HD  & P2F \\
    \midrule
    PU-Net   & 3.095  & 20.468 & 2.875
             & 4.194  & 21.651 & 3.682
             & 3.622  & 25.043 & 4.717 & 8.25 \\
    iPUNet   & 2.215 & 14.163 & 2.247  & - & - & - & - & - & - & - \\
    MPU      & 1.492 & 14.574 & 1.361
             & 2.320 & 15.937 & 2.085
             & 2.860 & 20.772 & 4.098 & 8.17 \\
    PU-GCN   & 1.454 & 14.438 & 1.323
             & 2.364 & 17.232 & 2.118
             & 2.675 & 18.529 & 3.975 & 8.21 \\
    PU-GAN   & 1.318 & 11.366 & 1.230
             & 2.005 & 11.744 & 1.893
             & 2.443 & 17.182 & 3.811 & 9.19 \\
    PU-CRN   & \cellcolor{orange!30}{0.845} & 5.741
             & \cellcolor{orange!30}{0.549}
             & 1.261  & 8.717 & 1.050
             & \cellcolor{orange!30}{2.241} & 9.659
             & \cellcolor{orange!30}{3.538} & 8.37 \\
    NePs     & 1.100 & 16.809 & 0.778
             & \cellcolor{red!30}{0.943}  & 10.123  & \cellcolor{red!30}{0.616} & - & - & - & 10.85\\
    Grad-PU  & 1.150 & \cellcolor{orange!30}{2.318} & {0.690}
             & 1.346 & \cellcolor{orange!30}{2.440} & {1.048}
             & 3.997 & \cellcolor{orange!30}{4.627} & {3.779}
             & \cellcolor{orange!30}{6.58} \\
    Ours    & \cellcolor{red!30}{0.801}    & \cellcolor{red!30}{1.044}
             & \cellcolor{red!30}{0.425}
             & \cellcolor{orange!30}{1.004} & \cellcolor{red!30}{1.869}
             & \cellcolor{orange!30}{0.768}
             & \cellcolor{red!30}{1.818}    & \cellcolor{red!30}{3.306}
             & \cellcolor{red!30}{3.458} & \cellcolor{red!30}{0.14} \\
    \bottomrule
    \end{tabular}
    \label{tab:UpsamplingQuan}
\end{table*}

\begin{table*}[htbp]
    \centering
    \setlength{\tabcolsep}{5.5pt}
    \caption{Quantitative comparisons for point cloud cleaning on PU-Net dataset in terms of CD ($\times$$10^{-5}$) and P2F ($\times$$10^{-3}$), lower is better. The percentage indicate the level of noise. Time is measured in seconds. Best results are highlighted as \colorbox{red!30}{first}, \colorbox{orange!30}{second}.}
    \begin{tabular}{l|rr|rr|rr|r|rr|rr|rr|r}
    \toprule
    \multirow{3}{*}{Models} & \multicolumn{7}{c|}{10k} & \multicolumn{7}{c}{50k} \\
    \cmidrule{2-15} &
    \multicolumn{2}{c|}{1$\%$}  & \multicolumn{2}{c|}{2$\%$} & \multicolumn{2}{c|}{2.5$\%$}& \multirow{2}{*}{Time} &
    \multicolumn{2}{c|}{1$\%$}  & \multicolumn{2}{c|}{2$\%$} & \multicolumn{2}{c|}{2.5$\%$}& \multirow{2}{*}{Time} \\
    \cmidrule{2-7}
    \cmidrule{9-14}
            &  CD   &  P2F  &  CD  &  P2F  &  CD  & P2F &  & CD &  P2F  &  CD   &  P2F &  CD  & P2F & \\
    \midrule
    PointCleanNet   & 11.22 & 2.45 & 14.71 & 3.81 & 16.75 & 4.87 & 3.24 & 2.81 & 1.54 & 4.13 & 3.35 & 5.17 & 4.93 & 18.96\\
    ScoreDenoise   & 9.32 & 2.09 & 13.40 & 3.60 & 16.44 & 3.86 & \cellcolor{orange!30}{1.27} & 2.58  & 1.49 & 3.76 & 3.19 & 4.89 & 3.04 & 12.21 \\
    PointFilter   & 9.30 & 1.85 & 13.82 & 2.90 & 15.35 & 3.43 & 4.33 & 2.74 & 1.10 & 3.90 & 1.82 & 4.18 & 2.11 & 16.19 \\
    PCDNF   & 9.12 & 1.72 & 13.72 &  2.75 &  15.04 & 3.17 & 5.14 & 2.52 & 1.00 & 3.48 & 1.86 & 3.89 & 2.31 & 23.75 \\
    IterativePFN    & \cellcolor{orange!30}{7.75} & \cellcolor{red!30}{1.29}
                    & \cellcolor{orange!30}{12.21}& \cellcolor{red!30}{2.13}
                    & \cellcolor{orange!30}{13.48}& \cellcolor{red!30}{2.58} & 5.19
                    & \cellcolor{orange!30}{2.42} & \cellcolor{red!30}{0.82}
                    & \cellcolor{orange!30}{3.19} & \cellcolor{red!30}{1.50}
                    & \cellcolor{orange!30}{3.78} & \cellcolor{orange!30}{2.11} & \cellcolor{orange!30}{11.22} \\
    Ours & \cellcolor{red!30}{7.59}    & \cellcolor{orange!30}{1.52}           & \cellcolor{red!30}{9.62}    & \cellcolor{orange!30}{2.49}
          & \cellcolor{red!30}{10.30}   & \cellcolor{orange!30}{2.95}
          & \cellcolor{red!30}{0.13}
          & \cellcolor{red!30}{1.78}    & \cellcolor{orange!30}{0.94}
          & \cellcolor{red!30}{2.61}    & \cellcolor{orange!30}{1.51}
          & \cellcolor{red!30}{3.13}    & \cellcolor{red!30}{1.95}
          & \cellcolor{red!30}{0.13} \\
    \bottomrule
    \end{tabular}
    \label{tab:cleaningQuan}
\end{table*}

\subsection{Comparisons} \label{sec:comparisons}
\paragraph{Point Cloud Upsampling.}
We compared our method with 7 state-of-the-art methods: PU-Net, MPU, PU-GAN, PU-GCN, PU-CRN, Grad-PU, Neural Point (NePs). These methods took patches cropped from a whole sparse point cloud as input and predicted a dense point cloud with a constant number of output points. In contrast, our method represented the point cloud as an adaptive octree structure, resulting in a variable number of output points. To ensure a fair comparison, we adjusted our output point cloud to make the average number of whole testing set approximate other methods while evaluating CD, HD, and P2F.

\paragraph{Point Cloud Cleaning.}
We conducted comparisons with several state-of-the-art point cloud cleaning methods: PointCleanNet, PointFilter, ScoreDenoise, PCDNF, IterativePFN. Following the majority of existing methods~\cite{de2023iterativepfn,luo2021score,wang2023random,luo2020differentiable}, we employed CD and P2F for quantitative comparisons. To ensure a fair comparison, we downsampled our output point cloud to the same resolution as the other methods before evaluating the results.

\begin{figure*}[htb]
    \centering
   \includegraphics[width=0.98\linewidth]{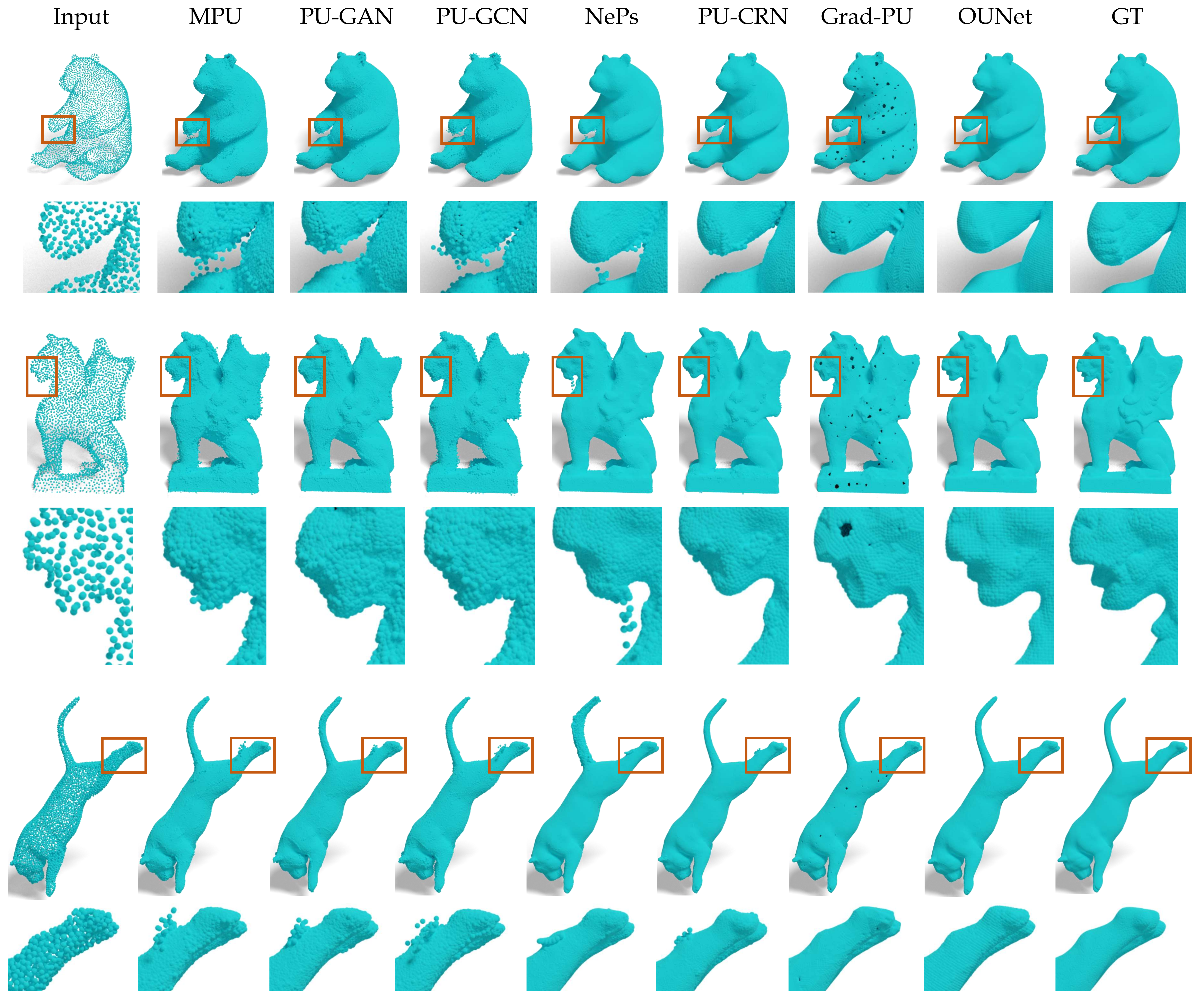}
    \caption{Qualitative upsampling results on the PU-GAN dataset. We provide a zoom-in view of the orange box for clearer observation. Outputs of our model are visually better, with fewer outliers, more fine-grained and sharp geometry and more uniform distribution.}
   \label{fig:PUGANimage}
\end{figure*}

\subsection{Results} \label{sec:results}
\begin{figure*}[htbp]
   \centering
   \includegraphics[width=0.89\linewidth]{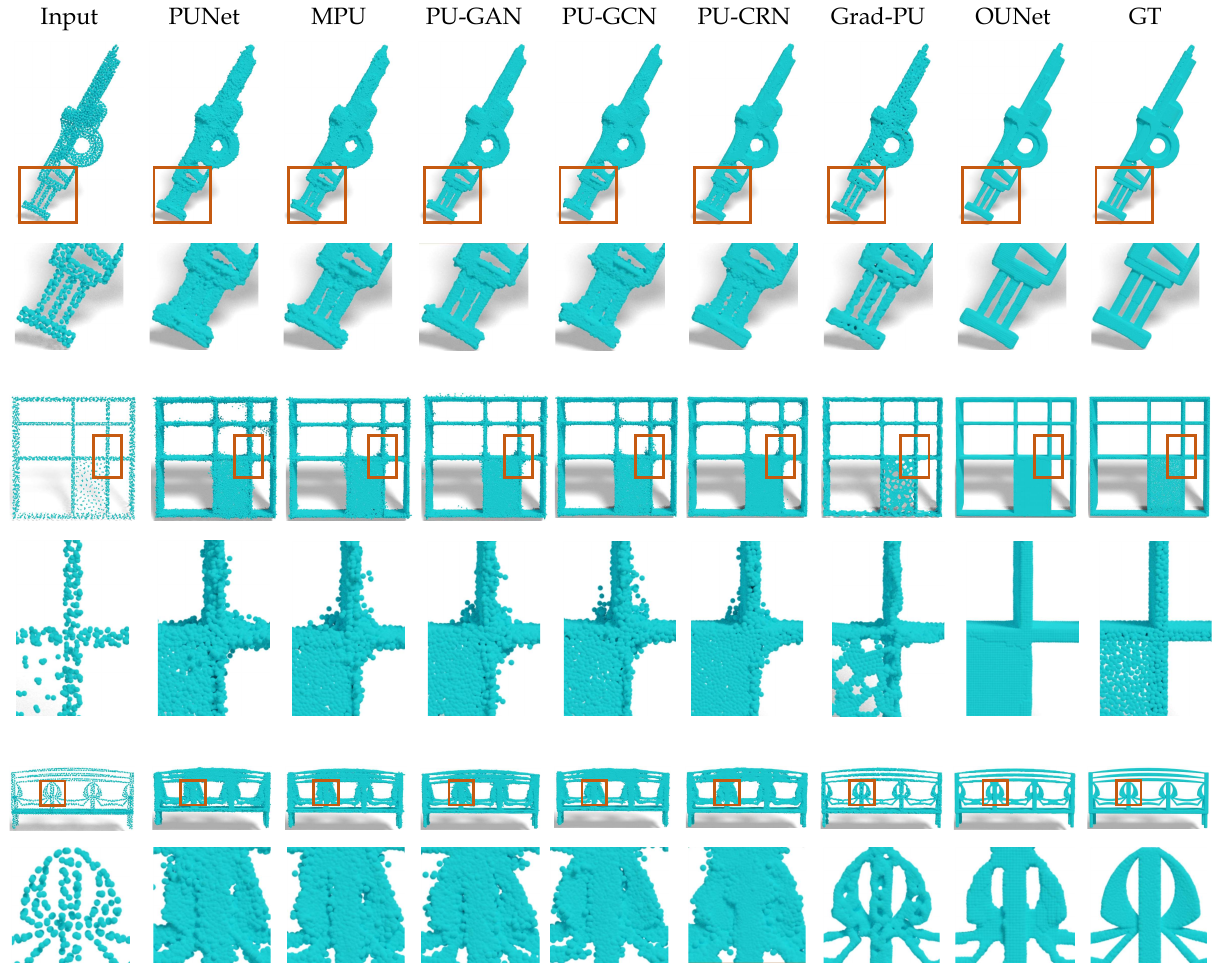}
   \caption{Qualitative upsampling results on the PU1K dataset.}
   \label{fig:PU1kimage}
\end{figure*}

\paragraph{Point Cloud Upsampling.}
In Tab.~\ref{tab:UpsamplingQuan}, we presented results on the PU-GAN, Sketchfab, and PU1K datasets. Our method achieved the best performance on both the PU-GAN and PU1K datasets across all three metrics. Additionally, it showed greater improvement over existing methods as the number of training samples increases, as evidenced by the PU1K dataset, which contains significantly more training meshes (900) than PU-GAN and Sketchfab.
Visual results for the PU-GAN and PU1K datasets were shown in Fig.~\ref{fig:PUGANimage} and \ref{fig:PU1kimage}, respectively. Odd rows displayed the entire upsampled shapes, while even rows provided zoomed-in views for clearer comparison. As shown, our method achieved the highest quality, with fewer outliers, sharper geometry, and a more uniform distribution.

\begin{figure*}[htbp]
   \centering
   \includegraphics[width=0.87\linewidth]{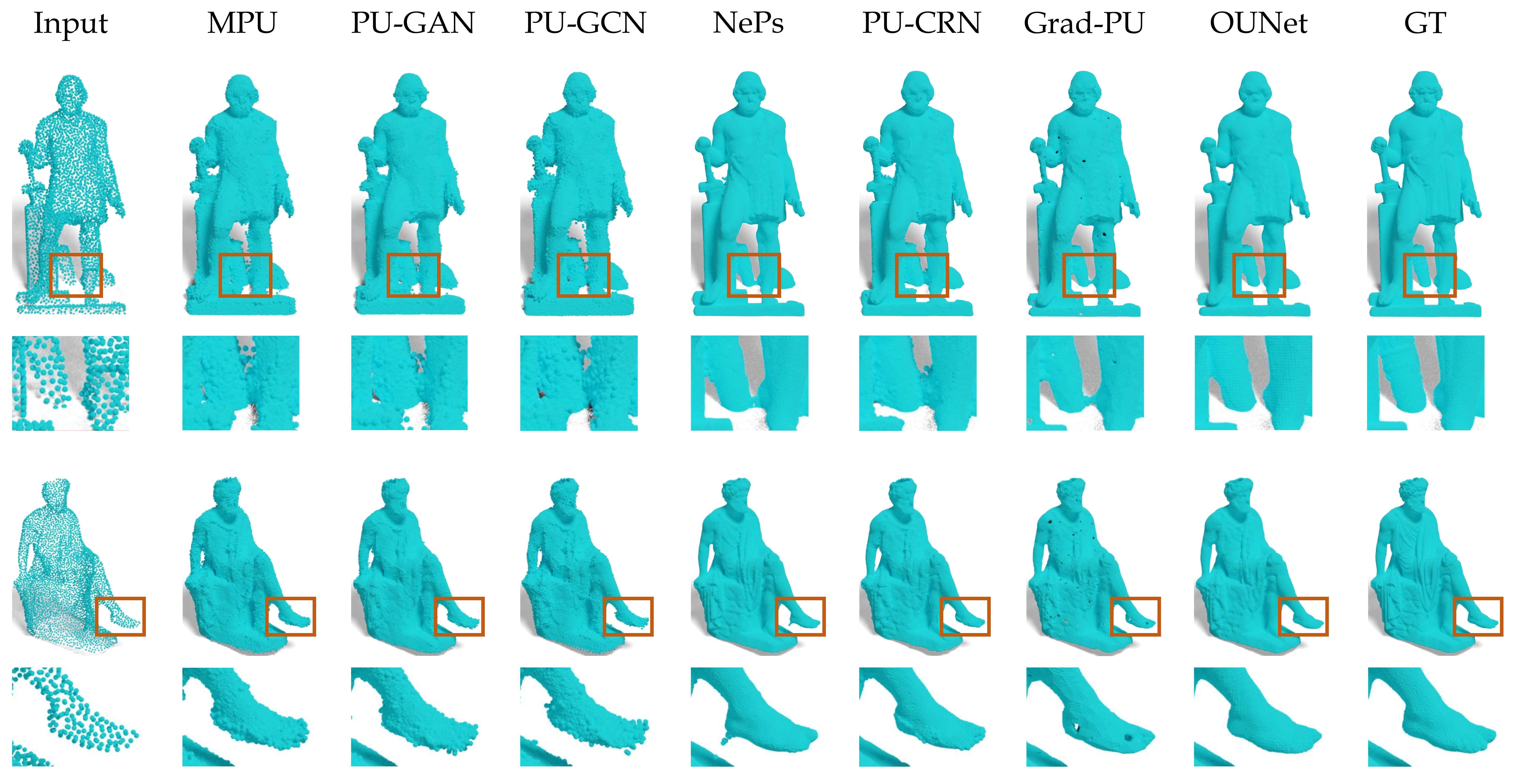}
   \caption{Qualitative upsampling results on the Sketchfab dataset.}
   \label{fig:Sketchfabimage}
\end{figure*}

For Sketchfab, while our CD and P2F were slightly higher than Neural Points, we achieved the lowest HD. As seen in Fig.~\ref{fig:Sketchfabimage}, our method also performed better visually, with significantly fewer outliers or holes.
The slightly worse CD and P2F may result from:
(\romannumeral1) Limited training data: Sketchfab only provides 90 meshes for training, which may be insufficient for our method.
In contrast, Neural Points splits the whole point cloud into 1000 patches during training, resulting in a total of 90,000 training samples.
(\romannumeral2) Less input information required: Neural Points utilizes both coordinates and normals of points. On the contrary, our method does not rely on normals.

Finally, we compared inference times for generating the entire point cloud, as shown in the last column of Tab.~\ref{tab:UpsamplingQuan}. Without requiring time-consuming pre- and post-processing steps, our method achieved significantly faster inference speeds. We also presented the CD-Time diagram for the PU-GAN dataset in Fig.~\ref{fig:speed-up}, further highlighting the efficiency of our method.

\paragraph{Synthetic Data Cleaning.}
The quantitative results for the PU-Net dataset were summarized in Tab.~\ref{tab:cleaningQuan}, covering noise levels of 1$\%$, 2$\%$ and 2.5$\%$, as well as resolutions of 10,000 and 50,000 points.
Our method consistently outperformed most existing techniques and performed comparably to IterativePFN. While our approach generally resulted in lower CD but slightly higher P2F, it's worth noting that IterativePFN is more complex and runs significantly slower. As shown in Fig.~\ref{fig:speed-10} and \ref{fig:speed-50}, our method demonstrated remarkable efficiencyt, running over 10 and 53 times faster than others for input point clouds with $10k$ and $50k$ points, respectively.

\begin{figure*}[htbp]
   \centering
   \includegraphics[width=0.9\linewidth]{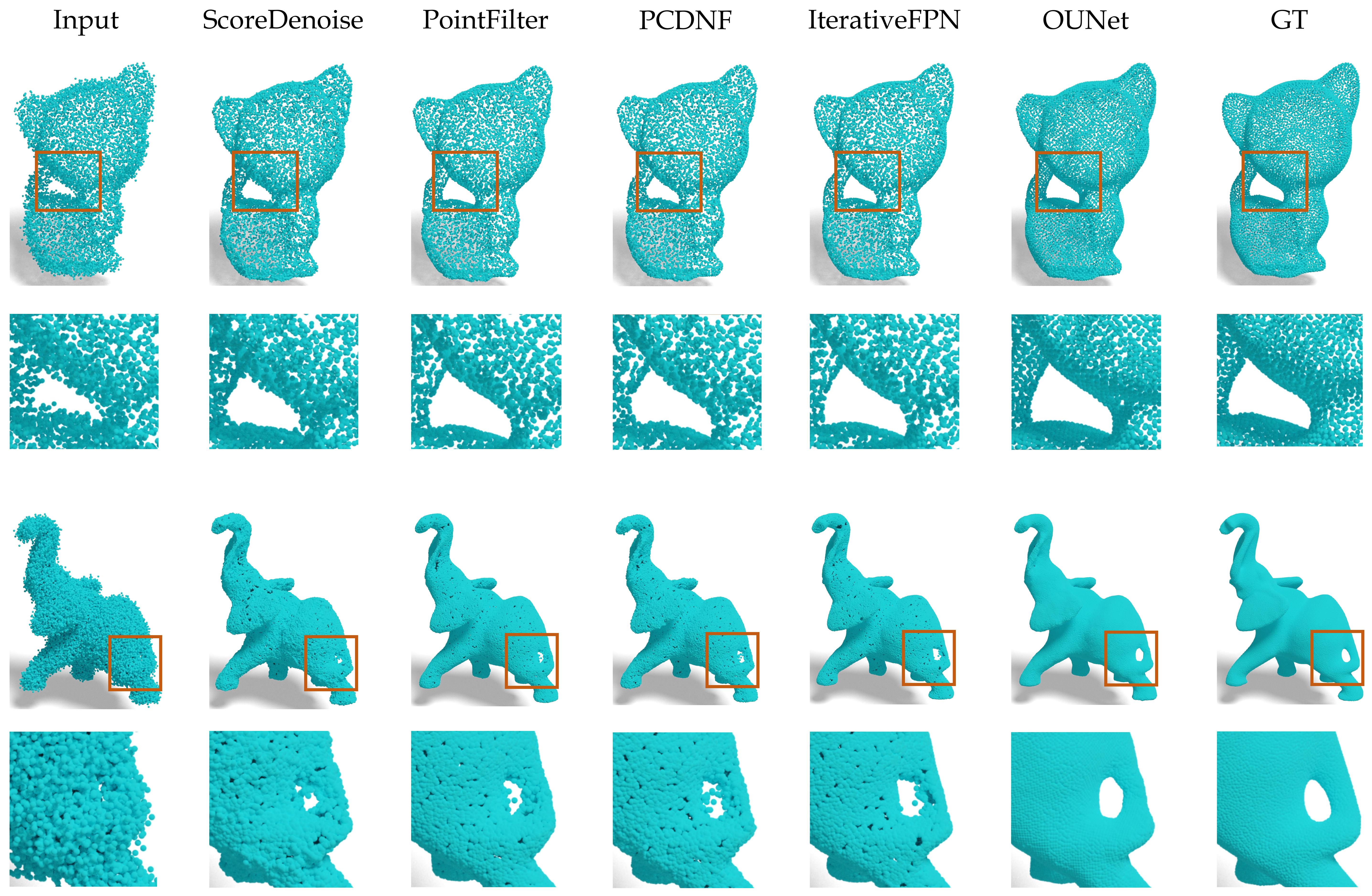}
   \caption{Qualitative cleaning results on the PU-Net test dataset.}
   \label{fig:cleanimage}
\end{figure*}

Fig.~\ref{fig:cleanimage} presented the qualitative results for inputs with resolutions of $10k$ and $50k$ in the first and last two rows, respectively, both at a noise level of 0.02. As shown, our method produced more uniform distributions and smoother boundaries across most test samples.

\begin{figure*}[t]
    \centering
    \begin{subfigure}[b]{0.445\textwidth}
        \centering
        \includegraphics[width=0.95\textwidth]{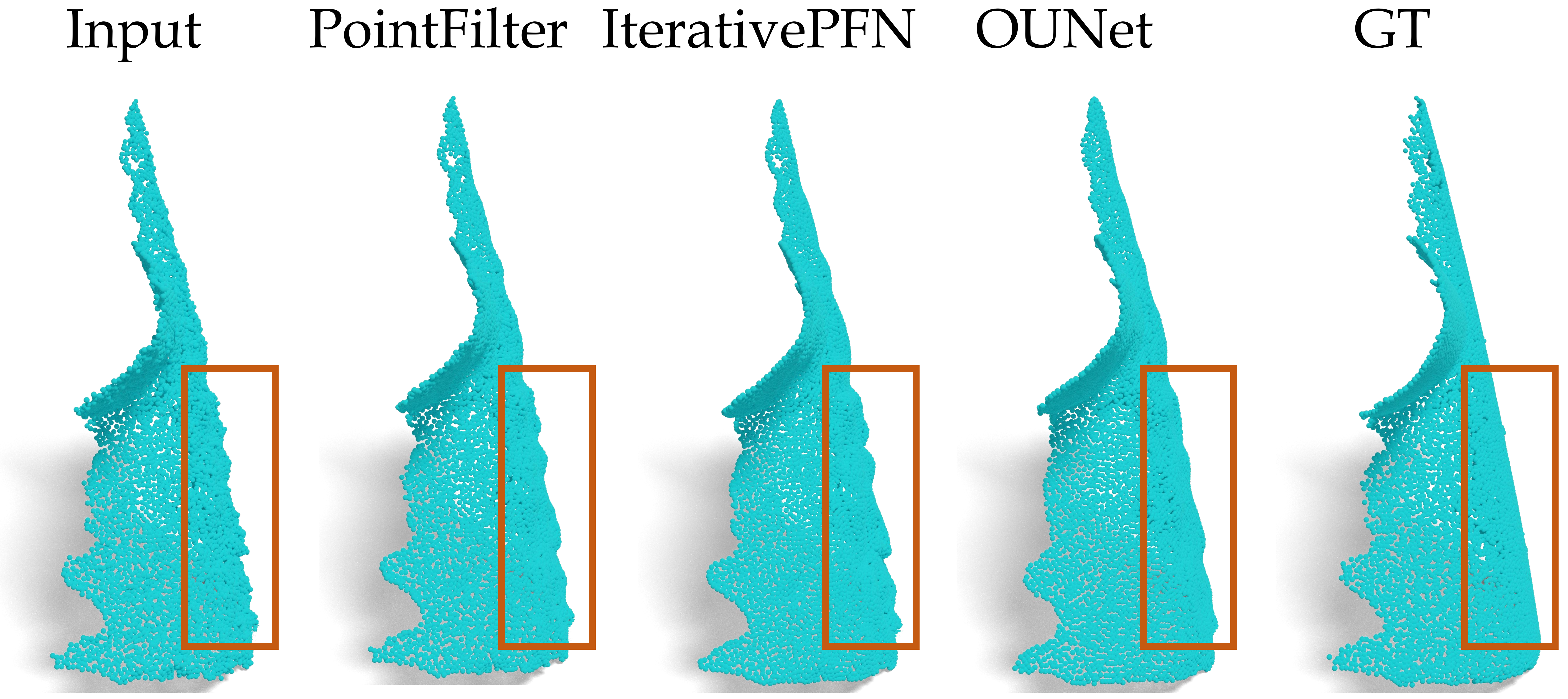}
        \caption{Kinect v1}
        \label{fig:kinect_v1}
    \end{subfigure}
    \hfill
    \begin{subfigure}[b]{0.455\textwidth}
        \centering
        \includegraphics[width=0.85\textwidth]{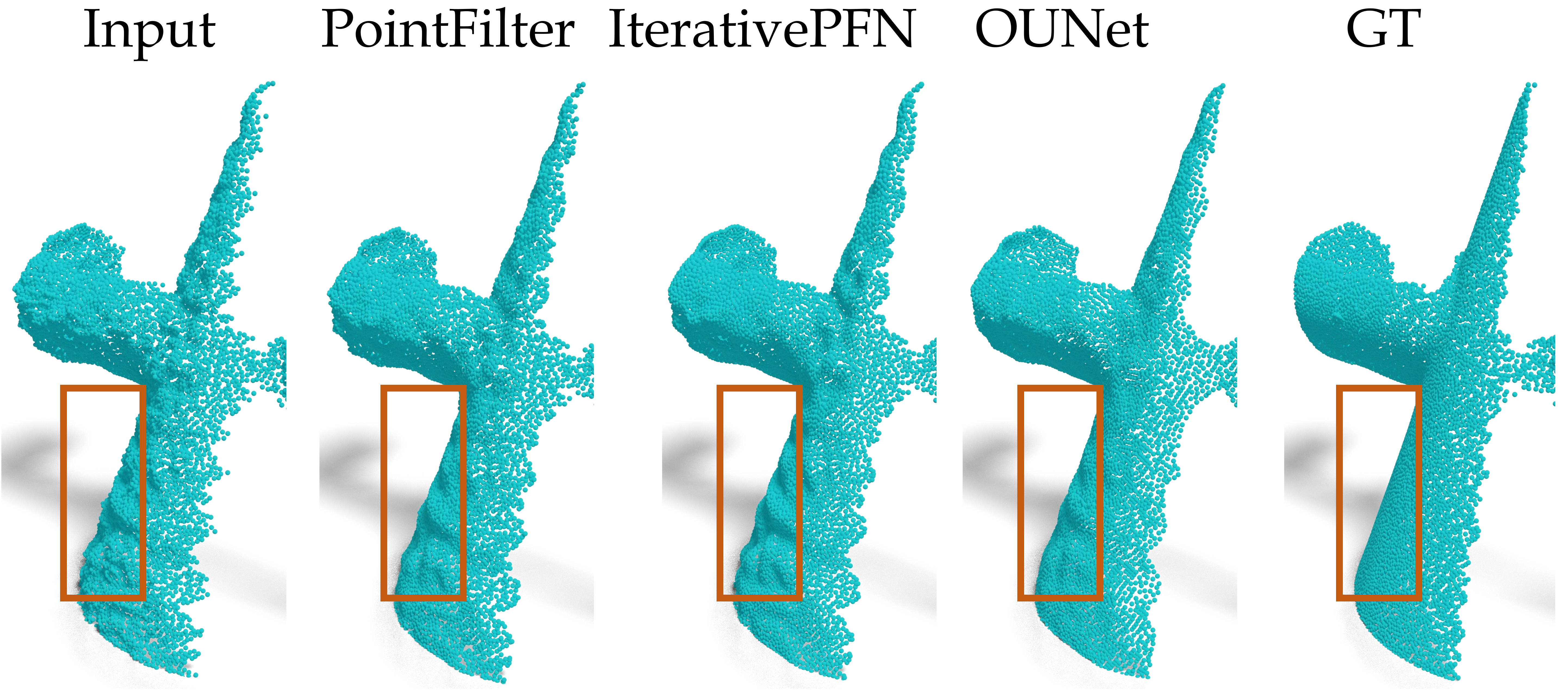}
        \caption{Kinect v2}
        \label{fig:kinect_v2}
    \end{subfigure}
    \caption{Qualitative evaluation on the Kinect datasets. The results from OUNet achieve the neatest surface among the compared methods.}
    \label{fig:real-world}
\end{figure*}

\begin{figure*}[h!]
    \centering
    \begin{subfigure}[b]{0.44\textwidth}
        \centering
        \includegraphics[width=0.95\textwidth]{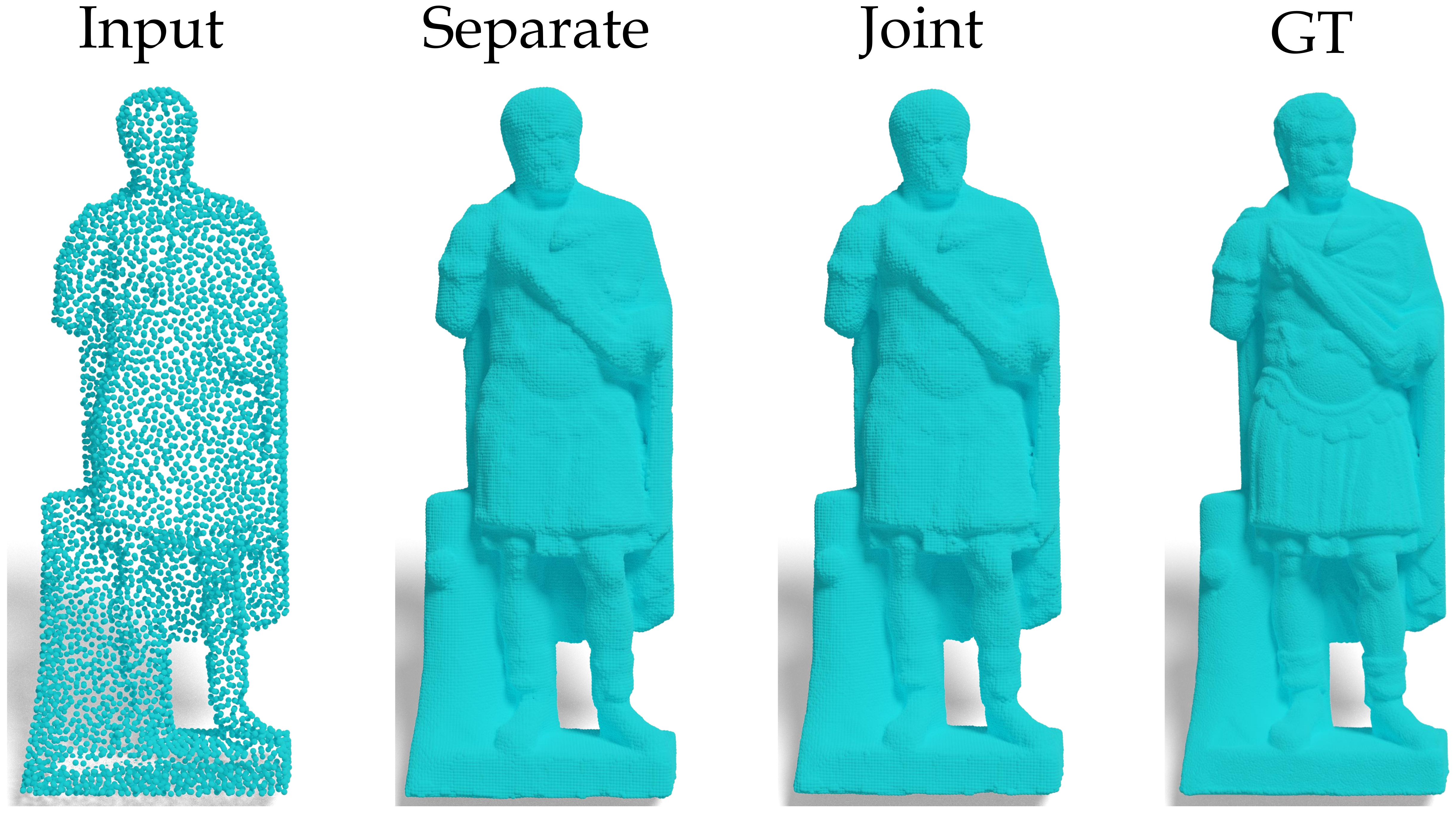}
        \caption{Upsampling}
        \label{fig:joint-upsampling}
    \end{subfigure}
    \hfill
    \begin{subfigure}[b]{0.48\textwidth}
        \centering
        \includegraphics[width=0.95\textwidth]{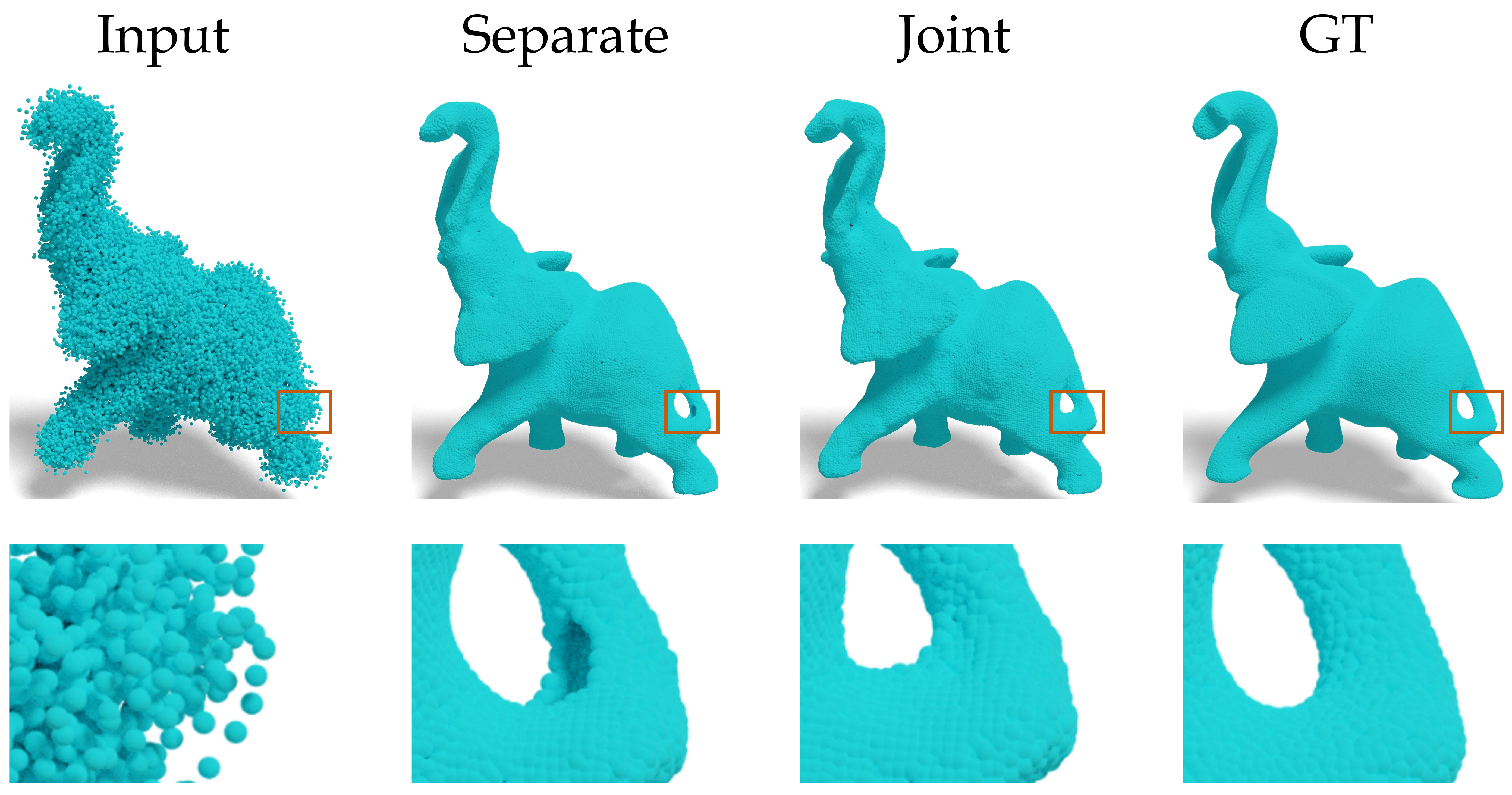}
        \caption{Cleaning with 50k points and 2.5$\%$ noise.}
        \label{fig:joint-cleaning}
    \end{subfigure}
    \caption{Visual comparisons between joint and separete training. (a) shows a case in PU-GAN dataset. (b) demonstrates the point cloud cleaning with resolution of 50k and 2.5$\%$ noise.}
    
    \label{fig:joint}
\end{figure*}

\begin{table*}[htbp]
    \caption{Numerical results for separate training and ablation runs, in terms of CD ($\times$$10^{-5}$), HD ($\times$$10^{-4}$), P2F ($\times$$10^{-3}$).
    The results of cleaning experiments are from point cloud with resolution of 50k. Best results are highlighted using bold.}
    \label{tab:ablation}
    \centering
    \setlength{\tabcolsep}{6.7pt}
    \begin{tabular}{ccc|rr|rr|rr|rr|rr|rr}
    \toprule
    \multirow{2}{*}{GN}         & \multirow{2}{*}{Non-Patch} & \multirow{2}{*}{Joint} &
    \multicolumn{2}{c|}{PU-GAN} & \multicolumn{2}{c|}{Sketchfab} & \multicolumn{2}{c|}{PU1K} & \multicolumn{2}{c|}{Noise $1\%$ } & \multicolumn{2}{c|}{Noise $2\%$ } & \multicolumn{2}{c}{Noise $2.5\%$ } \\
    \cmidrule{4-15}
    &   &  &
    CD   & P2F  & CD   & P2F  & CD   & P2F   &  CD  & P2F  &  CD  &  P2F & CD   & P2F \\
    \midrule
    \ding{55} & \ding{51} & \ding{51} &
    0.81 & 0.47 & 1.01 & 0.89 & 2.64 & \textbf{3.45} & 4.81 & 4.85 & 7.25 & 8.17 & 8.40 & 10.09 \\
    \ding{51} & \ding{55} & \ding{51} &
    0.91 & 0.45 & 1.04 & 0.84 & \textbf{1.72} & 3.76 & \textbf{1.74} & 0.98 & 2.50 & 2.10 & 2.94 & 2.84 \\
    \ding{51} & \ding{51} & \ding{55} &
    0.84 & 0.43 & \textbf{0.97} & \textbf{0.72} & 2.03  & 3.46 & 1.78 & 0.98 & \textbf{2.43} & 1.56 & \textbf{2.93} & 1.96 \\
    \ding{51} & \ding{51} & \ding{51} &
    \textbf{0.80} & \textbf{0.43}     & 1.00 & 0.77 & 1.82 & 3.46 & 1.78 & \textbf{0.94} & 2.61 & \textbf{1.51} & 3.13  & \textbf{1.95} \\
    \bottomrule
    \end{tabular}
\end{table*}

\begin{table}[htbp]
    \center
    \setlength{\tabcolsep}{9.5pt}
    \caption{Quantitative comparisons on real-scaned datasets: Kinect v1 and v2 in terms of CD ($\times$$10^{-5}$), HD ($\times$$10^{-4}$), P2F ($\times$$10^{-3}$), lower is better. Best results are highlighted as bold.}
    \begin{tabular}{l|rr|rr}
    \toprule
    \multirow{2}{*}{Models} & \multicolumn{2}{c|}{Kinect v1} & \multicolumn{2}{c}{Kinect v2} \\
    \cmidrule{2-5}
    &  CD  &  P2F  &  CD  &  P2F \\
    \midrule
    PointFilter   & 5.322
                  & 2.804
                  & 5.896
                  & 3.544 \\
    IterativePFN  & 5.539 & 2.942 & 5.956 & 3.612 \\
    Ours          & \textbf{5.169}
                  & \textbf{2.745}
                  & \textbf{5.737}
                  & \textbf{3.532} \\
    \bottomrule
    \end{tabular}
    \label{tab:Kinect}
    \end{table}

\paragraph{Real-world Scans Cleaning.}
In real-world scenarios, ground-truth data is often unavailable. Therefore, we leveraged the parameters trained on synthetic datasets to infer the cleaning results for the Kinect v1 and v2 datasets. As shown in Tab.~\ref{tab:Kinect}, our method outperformed others in terms of the CD and P2F metrics. The qualitative comparisons in Fig.~\ref{fig:real-world} further demonstrated that OUNet produced the neatest surfaces, showcasing the strong generalization capability of our method.

\subsection{Ablation Studies} \label{sec:ablation}
\paragraph{Separate Upsampling and Cleaning.} \label{sec:separate}
We trained the OUNet separatly using four aforementaioned datasets in Sec.~\ref{sec:dataset}, and compared joint and separate training for upsampling and cleaning from both numerical and visual perspectives. As shown in Tab.~\ref{tab:ablation}, the results from a single network trained jointly were comparable to and even better than those from training separate networks for each task. This joint scheme allowed for flexible task execution, requiring only one model to be trained and saved, which benefits computational and storage resources.
Furthermore, Fig.~\ref{fig:joint} demonstrated that the cleaning results from joint training achieving a more complete point cloud. We supposed that the small size of the cleaning dataset limits the network's ability when trained separately, and joint training can enrich the training dataset.

\vspace{-2mm}
\paragraph{Group Normalization.}
GN instead of BN is crucial for our network. BN mixed the upsampling and cleaning feature maps, leading to a sharp decline in performance, while GN only focusing on a case addressed this issue. As shown in Tab.~\ref{tab:ablation}, the CD and P2F metrics in cleaning experiments with GN were significantly smaller than those using BN.

\paragraph{Non-patch Style.}
Processing the entire point cloud, as opposed to processing patches, facilitates the extraction of global features and improves performance, particularly in the presence of significant noise. Following the split implementations of Grad-PU and IterativePFN for upsampling and cleaning dataset respectively, we trained the OUNet using a patch-based strategy, then stitched the resulting patches through concatenation and FPS algorithm. As shown in Tab.~\ref{tab:ablation}, we compared these results with those obtained from the non-patch-based approach, and the results demonstrated that non-patch strategy surpassed patch-based method in almost all the evaluation criterion. Notably, patch-based strategy requires time-consuming pre- and post-processing, leading to waste lots of computation resources.

%% file: src/conclusion.tex
\section{Conclusion} \label{sec:conclusion}
This paper introduces a joint training framework for point cloud upsampling and cleaning based on an octree-based U-Net. 
Unlike previous methods, which rely on PointNet or DGCNN for patch-wise processing, our approach operates on the entire point cloud, simplifying the network architecture and reducing pre- and post-processing steps. 
By replacing batch normalization with octree-based group normalization, we achieve a more efficient training process without adding new components. 
This modification enhances the performance of both tasks, offering an accessible and effective baseline for future research.

The limitation and potential future work are as follows:

\textit{Denser Ground-truth}:  
    One limitation of our method is the need to sample denser point clouds for ground-truth octree construction, though not for the input point cloud. This is necessary due to the downsampling required when building the octree from the point cloud.

\textit{Diverse Outputs}:
    Our focus has been on using OUNet for the prediction of dense or clean point clouds. However, these tasks often correspond to multiple possible results when upsampling or cleaning a low-quality point cloud. Thus, exploring the capability to generate diverse high-quality results for these tasks would be an interesting direction.

%% file: src/acknowledgements.tex
\subsection*{Acknowledgements}
We thank the anonymous reviewers for their invaluable suggestions.  This work is supported by Beijing Natural Science Foundation No. 4244081.